\documentclass[10pt,twocolumn,letterpaper]{article}

\usepackage{multirow}
\usepackage[pagenumbers]{cvpr} 

\usepackage{multibib}
\newcites{supp}{References}

\definecolor{cvprblue}{rgb}{0.21,0.49,0.74}
\usepackage[pagebackref,breaklinks,colorlinks,allcolors=cvprblue]{hyperref}
\usepackage{bbding}
\usepackage{pifont}
\usepackage{color,soul}

\usepackage{algorithm}
\usepackage{algorithmic}

\def\paperID{9303} 
\def\confName{CVPR}
\def\confYear{2025}

\title{DiffSLT: Enhancing Diversity in Sign Language Translation via Diffusion Model}

\author{\normalsize
JiHwan Moon$^{1*}$\hspace{1em}
Jihoon Park$^{2*}$\hspace{1em}
Jungeun Kim$^{1*}$\hspace{1em}
Jongseong Bae$^{1*}$\hspace{1em}
Hyeongwoo Jeon$^{2}$\hspace{1em}
Ha Young Kim$^{2\dagger}$\\
\normalsize
$^{1}$ Department of Artificial Intelligence, Yonsei University\\
\normalsize
$^{2}$ Graduate School of Information, Yonsei University\\
{\tt\small \{hoho, zzang9jihoon, jekim5418, js.bae, hyeong1204, hayoung.kim\}@yonsei.ac.kr}
}

\begin{document}

\definecolor{yellow}{HTML}{f9f871}
\definecolor{blue}{HTML}{B7D9DE}
\definecolor{pink}{HTML}{EED1F8}

\definecolor{darkblue}{HTML}{000066}
\DeclareRobustCommand{\hlblue}[1]{{\sethlcolor{blue}\hl{#1}}}
\DeclareRobustCommand{\hlpink}[1]{{\sethlcolor{pink}\hl{#1}}}
\DeclareRobustCommand{\hlyellow}[1]{{\sethlcolor{yellow}\hl{#1}}}

\twocolumn[{
\maketitle
\begin{center}
  \centering
    \captionsetup{type=figure}
    \includegraphics[width=1.\textwidth]{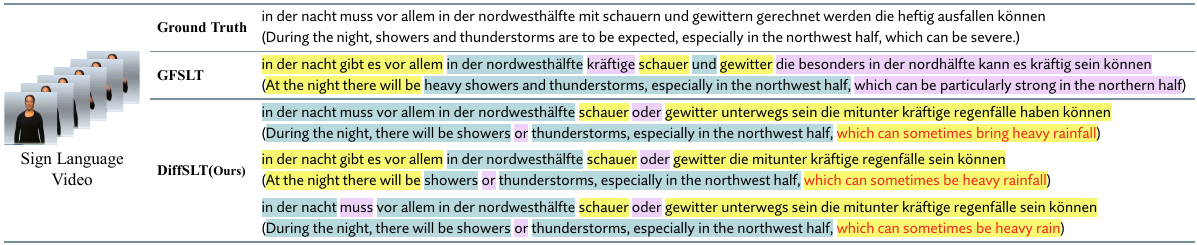}
    \captionof{figure}{Translation results on the PHOENIX14T~\cite{Camgoz_2018_CVPR}. DiffSLT generates multiple high-quality translations that are both diverse and accurate, selecting the sentence closest to the ground truth. In contrast, existing methods produce a single translation for a sign language video.
    \hlblue{Blue} indicates a correct translation, 
    \hlpink{purple} represents an incorrect translation, and 
    \hlyellow{yellow} denotes cases where different words with the same meaning or alternative word choices appear across translation candidates, all conveying the same underlying meaning.
    }
    \label{fig:concept}
    \vspace{0.1em}      
\end{center}}]

\begin{abstract}
Sign language translation (SLT) is challenging, as it involves converting sign language videos into natural language.
Previous studies have prioritized accuracy over diversity.
However, diversity is crucial for handling lexical and syntactic ambiguities in machine translation, suggesting it could similarly benefit SLT.
In this work, we propose DiffSLT, a novel gloss-free SLT framework that leverages a diffusion model, enabling diverse translations while preserving sign language semantics.
DiffSLT transforms random noise into the target latent representation, conditioned on the visual features of input video.
To enhance visual conditioning, we design Guidance Fusion Module, which fully utilizes the multi-level spatiotemporal information of the visual features.
We also introduce DiffSLT-P, a DiffSLT variant that conditions on pseudo-glosses and visual features, providing key textual guidance and reducing the modality gap. 
As a result, DiffSLT and DiffSLT-P significantly improve diversity over previous gloss-free SLT methods and achieve state-of-the-art performance on two SLT datasets, thereby markedly improving translation quality. 
Project page: \texttt{\href{https://diffslt.github.io/}{https://diffslt.github.io/}}.
\end{abstract}
\vspace{-.2cm}    

\vspace*{\fill} 
\noindent\rule{7cm}{0.3pt}
\\
\vspace{0.4em}      
{\footnotesize
\noindent\textsuperscript{*}Equal contribution;
\textsuperscript{\dag}Corresponding author.
}
\newpage 

\section{Introduction}
\label{sec:intro}
\begin{figure*}[!t]
    \centering
    \includegraphics[width=0.9\textwidth]{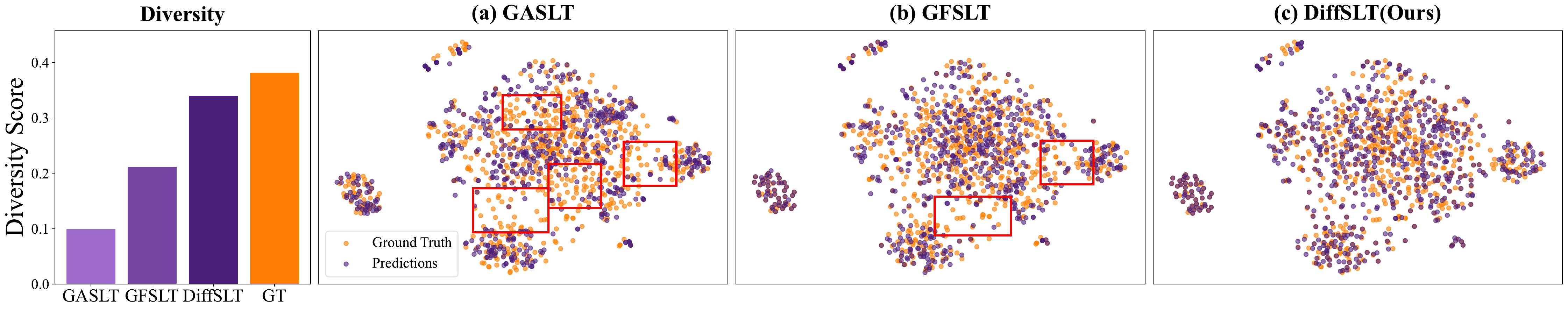}
    \caption{Comparison of diversity scores and distributions of translated spoken sentences.
    Previous SLT models exhibit relatively low diversity scores, with the predicted spoken sentences showing a distorted distribution in the text embedding space compared to the ground truth. In contrast, the proposed DiffSLT demonstrates a diversity score and distribution that closely resemble those of the ground truth.
    }
    \label{fig:observation}
    \vspace{-.4cm}
\end{figure*}

Sign language utilizes various visual cues—including hand movements, facial expressions, and mouth shapes—and is the primary communication method for deaf individuals~\cite{dreuw2007speech}.
Sign language translation (SLT) involves translating sign language video sequences into spoken sentences. This task is challenging due to data variability (\eg, variations in signer appearances and motions) and the modality gap between sign language and natural language~\cite{chen2022two}.

A typical sign language dataset~\cite{Camgoz_2018_CVPR,zhou2021improving} includes sign language video sequences, target spoken sentences, and textual annotations for sign language sequences (\ie, glosses).
Glosses provide an effective intermediate supervision due to their textual form and direct sequential alignment with sign language videos~\cite{chen2022two, zhao2021conditional}.
Several gloss-based models~\cite{chen2022two,yao2023sign,chen2022simple,zhang2023sltunet,jin2022prior,yin2021simulslt,voskou2021stochastic} have utilized this form of supervision from glosses.
However, gloss annotations require precise labeling by experts, which incurs significant human costs~\cite{zhou2023gloss}. To reduce dependency on gloss annotations, gloss-free models~\cite{li2020tspnet,yin2023gloss,zhou2023gloss,chen2024factorized,gong2024llms,wong2024sign2gpt} have been proposed to predict spoken sentences directly from sign video sequences, thereby enhancing the generalizability and scalability of SLT models.

While previous studies have focused on improving translation accuracy, they have not considered diversity a primary concern~\cite{gong2024llms,jiao2025visual}.
In neural machine translation (NMT), diverse translations are important for effectively handling lexical and syntactic ambiguities~\cite{wu2020generating, he2018sequence, sun2020generating}, suggesting that incorporating diversity could similarly improve the translation quality of SLT.
Our analysis shows that prior SLT models~\cite{yin2023gloss,zhou2023gloss} tend to lose representational diversity throughout the training phase. 
As described in Fig.~\ref{fig:observation}, The predicted distributions of prior 
SLT models are highly skewed in the text embedding space, indicating low diversity in their predictions. This lack of diversity can result in spoken sentences that do not adequately reflect the variability inherent in sign language, limiting their generalization to real-world use.

To tackle this problem, we propose a novel gloss-free SLT framework, DiffSLT, which leverages a diffusion model—a likelihood-based generative model—as a translator, enabling diverse translations while maintaining the semantics of sign language (Fig.~\ref{fig:concept}).
DiffSLT denoises random latent representations, transforming them into target sentence representations by using visual features from the input video as a condition.
To enhance diffusion conditioning, we design Guidance Fusion Module (GFM), which effectively integrates multi-level spatiotemporal features.
In addition, we introduce DiffSLT-P, a variant of DiffSLT that conditions on both visual features and pseudo-glosses predicted from visual features, to provide key textual guidance and reduce the modality gap.

Through extensive experiments, we demonstrate that the proposed DiffSLT and DiffSLT-P achieve state-of-the-art (SOTA) accuracy and effectively enhance the diversity of generated spoken sentences.
We find that both models significantly improve the distinctiveness of the generated phrases while preserving their underlying semantics within context.
The key contributions of our study are summarized as follows:
\begin{itemize}
    \item We propose DiffSLT, a novel gloss-free SLT framework for diverse translations. To the best of our knowledge, our method is the first to employ a diffusion model in SLT. 
    \item 
    To enhance diffusion conditioning of DiffSLT, we design a Guidance Fusion Module that effectively integrates multi-level spatiotemporal features.
    \item We propose an improved variant of DiffSLT, DiffSLT-P, which additionally utilizes pseudo-glosses for diffusion conditioning, providing key textual guidance during the denoising process and bridging the modality gap.
    \item Our extensive experiments demonstrate the superiority of the proposed DiffSLT and DiffSLT-P, significantly enhancing both diversity and accuracy, with both models achieving SOTA performance on standard benchmark datasets.
\end{itemize}
\begin{figure*}[!t]
    \centering
    \includegraphics[width=1\textwidth]{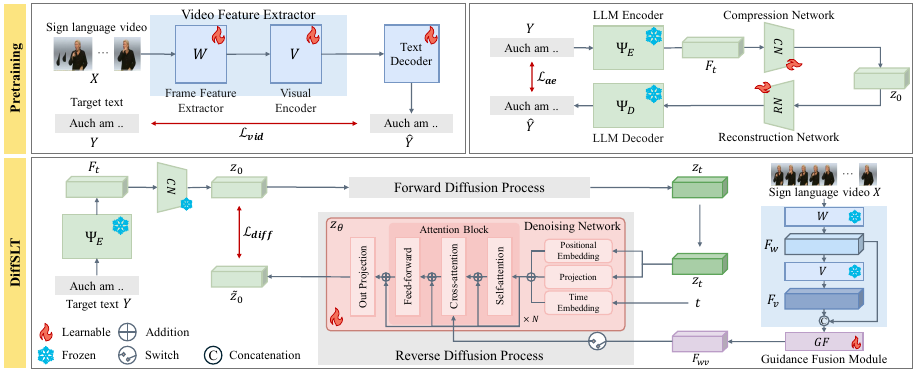}
    \caption{Overall training framework of DiffSLT.
    Our training process consists of two phases: pretraining for diffusion and diffusion training. 
    During pretraining, we extract text-aligned visual features and latent representations of spoken sentences. 
    In the diffusion training stage, our denoising network generates target sentence latent conditioned on the visual features obtained from pretraining.
    }
    \label{fig:DiffSLT overall}
\end{figure*}

\section{Related Works}
\noindent\textbf{Gloss-Free SLT.} Given the need for manual annotations and the limited expressiveness of gloss, recent research has advancements developing gloss-free models that eliminate the dependency on gloss annotations~\cite{wong2024sign2gpt,li2020tspnet,yin2023gloss,gong2024llms,chen2024factorized,zhou2023gloss,jiao2025visual}. 
Early gloss-free models recognized the presence of semantic continuity between adjacent frames and incorporated this observation into their model design~\cite{li2020tspnet,yin2023gloss}. Subsequent approaches aim to leverage the capabilities of LLMs by jointly training visual backbone and LLM decoder~\cite{gong2024llms,chen2024factorized}. 
Additionally, vision-language pretraining has been widely used to reduce the modality gap between sign language videos and spoken sentences, thus enhancing performance~\cite{zhou2023gloss,jiao2025visual}. 
However, these methods have predominantly emphasized accuracy improvements without adequately addressing the lexical and syntactic ambiguities that can arise in varying contexts~\cite{shen2024diverse}.

\noindent\textbf{Diffusion Model for Language Generation.} 
The diffusion model has demonstrated high-quality and diverse generation capabilities for continuous data~\cite{rombach2022high}, such as images~\cite{ho2020denoising, song2020denoising}, audios~\cite{schneider2023mo, huang2023noise2music}, and videos~\cite{jin2024pyramidal, zhou2024storydiffusion}. 
Motivated by these achievements, recent studies have proposed applying diffusion models to sequence-to-sequence (S2S) tasks within the natural language processing domain~\cite{yuan2022seqdiffuseq, lovelace2024latent, gong2022diffuseq, gao2022difformer, dieleman2022continuous, ye2023dinoiser}. 
These approaches generally transform discrete target text features into a continuous form and then train the model to restore the noised sample based on the target text distribution~\cite{li2022diffusion}. 
In this study, we begin by extending the application of diffusion models to SLT, thereby enabling the generation of diverse, high-quality translations.

\noindent\textbf{Diverse Translation.} In NMT, diversity is a primary challenge due to its critical role in managing the flexibility required for handling multiple possible interpretations of a single word or a sentence~\cite{nishida2024generating,li2016simple,wu2020generating,ruder2021xtreme,gong2022diffuseq}. 
Various approaches to enhance diversity have been proposed in NMT, including decoding strategies~\cite{nishida2024generating,li2016simple,gong2022diffuseq}, sampling methods~\cite{wu2020generating}, and data structuring~\cite{ruder2021xtreme}. 
In SLT, a concurrent study~\cite{shen2024diverse} tries to address the diversity problem, extending the existing datasets to include multiple candidate translations for a sign video and suggesting a strategy for choosing the optimal candidate.
Nonetheless, work in SLT has yet to address the diversity problem using a fundamentally generative model that inherently exhibits diversity in its generation.
In this study, we propose utilizing a diffusion model to achieve translations that are accurate and distinctive.
\section{Preliminary: Diffusion Model}
\label{Preliminary}
Diffusion model~\cite{ho2020denoising} is a generative model that progressively adds noise to an input sample and learns to reverse this process, gradually denoising the sample to match the target data distribution.
Given a data point $x_0$ and noise $\epsilon\sim \mathcal{N}(0, I)$, we can directly obtain the noised data point $x_t$ at timestep $t$ as follows:
\begin{equation}
  x_t = \sqrt{\Bar{\alpha}_t} \cdot x_0 + \sqrt{1-\Bar{\alpha}_t} \cdot \epsilon,
  \label{eq:prelim_dm1} 
\end{equation}
where $t$ is randomly sampled from the range [0, $T$], where $T$ is the total number of timesteps and $\Bar{\alpha}$ denotes the cumulative product of predefined timestep scheduling $\alpha$. 
To learn the denoising process from the noised data point, the objective is to predict the original input $x_0$ as follows:
\begin{equation}
  \mathcal{L} = \mathbb{E}_{t,\epsilon\sim\mathcal{N}(0, I), x_0\sim \mathcal{D}} \| x_0 - x_\theta(x_t, t) \|_2^2,
  \label{eq:prelim_dm2}
\end{equation}
where $\mathcal{D}$ represents the target data distribution, and $x_\theta$ denotes the prediction of input data. 
This denoising objective optimizes the model to remove noise from the noised sample at a specific timestep $t$, minimizing the error to recover the underlying data distribution $\mathcal{D}$. 

During inference, the diffusion model begins with a pure Gaussian noise sample $x_T \sim \mathcal{N}(0, I)$ and iteratively denoises it through each timestep $t = T, T-1, \dots, 1$ until reaching $t = 0$. 
At each step $t$, the model predicts the denoised sample $\tilde{x}_0$ based on the current noised sample $x_t$. 
We can then estimate the noised sample at the previous timestep $t-1$, denoted as $x_{t-1}$, based on $\tilde{x}_0$ by applying the noise addition in Eq.~\ref{eq:prelim_dm1}, where $x_0$ substituted with $\tilde{x}_0$.
In this progressive manner, each iteration moves the sample closer to the target data distribution $\mathcal{D}$, gradually denoising it until a final, realistic sample is generated.
\section{Method}
\label{sec:method}
In this section, we introduce DiffSLT, a gloss-free SLT framework designed to enable diverse translations by training a general distribution of spoken sentences.
The training process of DiffSLT consists of two stages: pretraining and main training.
First, in Sec.~\ref{sec:pretrain}, we describe the pretraining for DiffSLT, which aims to obtain text-aligned visual features and encode text embedding into the latent space for efficient latent diffusion.
Next, Sec.~\ref{sec:training_diffusion_model} outlines the training process for DiffSLT, where the model generates the target textual latent by conditioning on a combination of guidance from various supervision signals.
In Sec.~\ref{sec:inference}, we explain the complete SLT inference process for DiffSLT.
The overall training framework for DiffSLT is depicted in Fig.~\ref{fig:DiffSLT overall}.

\begin{figure}[!t]
    \centering
    \includegraphics[width=\linewidth]{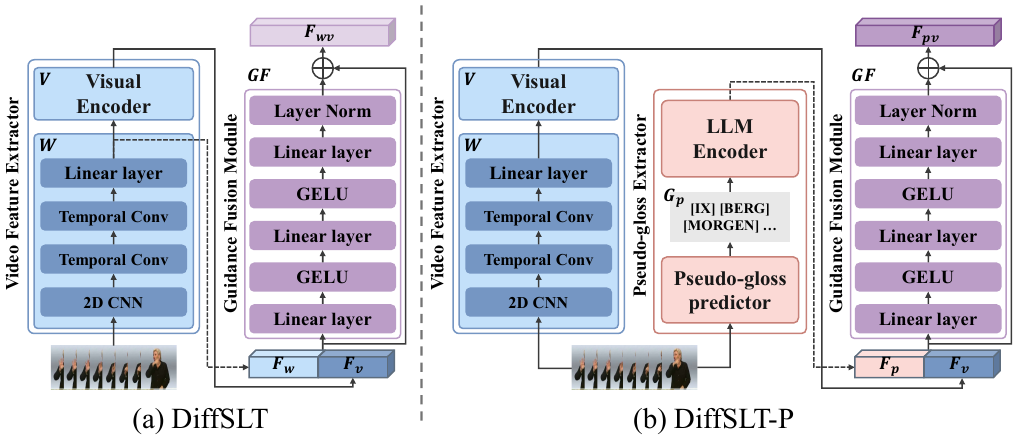}
    \caption{Illustration of Guidance Fusion Module. GFM provides unified representations by integrating two distinct levels of visual features, $F_v$ and $F_w$. In the weakly gloss-free setting (DiffSLT-P), $F_p$ replaces $F_w$ as the input to the GFM.
    }
    \label{fig:guidance_fusion}
\end{figure}

\subsection{Pretraining for DiffSLT}
\label{sec:pretrain}
\noindent\textbf{Video Feature Extractor.}
\label{sec:video_feature_extractor}
The modality gap between sign language and spoken sentences creates significant differences in feature distribution, complicating diffusion model training.
Thus, we employ a video feature extractor to derive text-aligned visual features from the video to enhance the conditioning of DiffSLT.
We pretrain this module to guide the diffusion model in generating textual latents.

The video feature extractor consists of two components: a frame feature extractor $W$ and a visual encoder $V$. 
First, $W$ extracts frame-level visual features $F_w \in \mathbb{R}^{B \times L_v \times D}$ from the input sign video $X \in \mathbb{R}^{B \times L_v}$, 
where $B$, $L_v$, and $D$ denote the mini-batch size, the number of frames, and the embedding dimension of each frame, respectively. 
Then, the Transformer-based~\cite{vaswani2017attention} visual encoder $V$ takes the frame features $F_w$ as input to generate spatiotemporal video features $F_v \in \mathbb{R}^{B \times L_v \times D}$, capturing global temporal dependencies across frames.

The extracted features $F_v$ are fed into a text decoder to predict the target text $Y \in \mathbb{R}^{B \times L_t}$, where $L_t$ is the sentence length.
The video feature extractor loss $\mathcal{L}_{vid}$ is computed using the cross-entropy loss as follows:

\begin{equation}
\mathcal{L}_{vid} = - \frac{1}{B} \sum_{i=1}^{B} \sum_{j=1}^{L_t} y_{ij} \log(\hat{y}_{ij}),
\label{eq:crossentropy}
\end{equation}
where $y_{ij}$ is the ground-truth label for the $j$-th token in the $i$-th batch of the target text $Y$, and $\hat{y}_{ij}$ is the corresponding logits predicted from the text decoder.
Note that the text decoder is not used during DiffSLT training.

\noindent\textbf{Compression and Reconstruction Network.}
\label{sec:compression_and_reconstruction_network}
To facilitate an efficient diffusion process, we build upon recent work~\cite{lovelace2024latent} by employing an autoencoder comprising a compression and a reconstruction network. This autoencoder encodes text embeddings of spoken sentences into fixed-length, low-dimensional latent representations.
We first obtain text embeddings, $F_t\in \mathbb{R}^{B \times L_t \times D}$, from the spoken sentence $Y$ using a frozen LLM encoder $\Psi_E$.
These embeddings are then compressed into a fixed-length latent representation, $z_0\in \mathbb{R}^{B \times l \times d}$, via the compression network $CN$, where $l$ and $d$ represent the fixed latent sequence length and the reduced embedding dimension, respectively. 
$z_0$ is then passed through a linear layer, followed by the addition of positional embeddings.
The reconstruction network $RN$ then reverses the compression process by reconstructing the text embeddings from the latent representation.
Finally, a frozen LLM decoder $\Psi_D$ generates the reconstructed sentence from the text embeddings. 
The compression-reconstruction networks are optimized using $\mathcal{L}_{ae}$, equivalent to the cross-entropy loss, similar to Eq.~\ref{eq:crossentropy}.

\noindent\textbf{Pseudo-Gloss Extractor.}
DiffSLT can optionally incorporate pseudo-glosses to form the variant DiffSLT-P.
Since gloss representations provide semantically aligned textual elements,
DiffSLT-P enhances the base model by introducing key textual guidance for the diffusion model, helping to reduce the modality gap between visual and textual features.
To extract the pseudo-glosses $G_p \in \mathbb{R}^{B \times L_p}$, we use an off-the-shelf gloss prediction model pretrained on the continuous sign language recognition task~\cite{hu2024corrnet+}, where $L_p$ is the pseudo-gloss length.
These pseudo-glosses are subsequently transformed into text embeddings $F_p \in \mathbb{R}^{B \times L_p \times D}$ using a pretrained, frozen LLM encoder. The gloss prediction model remains frozen during diffusion training.

\subsection{DiffSLT Training}
\label{sec:training_diffusion_model}
In this section, we introduce the DiffSLT training process. 
A brief summary of the training process is as follows: 
starting from a target spoken sentence, the LLM encoder and compression network transform the sentence into low-dimensional latent features.
Simultaneously, the video feature extractor generates text-aligned visual features, which are then passed to the GFM to combine multiple levels of visual information. 
These fused visual features serve as conditioning input for the diffusion module, which subsequently generates the target text latents.

\noindent\textbf{Guidance Fusion Module.}
\label{sec:guidance_fusion_module} 
The generation quality of the diffusion model is highly dependent on the conditioning information~\cite{bansal2023universal}. 
To address this, we propose Guidance Fusion Module $GF$.
This module effectively conveys rich spatiotemporal information by integrating multi-level visual features, thereby improving the generated text quality.
As illustrated in Fig.~\ref{fig:guidance_fusion} (a), $GF$ combines the frame feature $F_w$ and video feature $F_v$ to generate a unified representation $F_{wv} = GF(F_w\ \copyright\ F_v) \in \mathbb{R}^{B \times (L_v + L_v) \times D}$ where \copyright \  denotes concatenation along the length dimension.
This unified representation is projected into a latent space using a three-layer feed-forward network with GELU activations~\cite{hendrycks2016gaussian} and layer normalization~\cite{ba2016layer}. 
The GFM is jointly trained with the denoising network to enhance conditioning information for better generation quality.

For DiffSLT-P, as shown in Fig.~\ref{fig:guidance_fusion} (b), the textual pseudo-gloss representation $F_p$ replaces the frame feature $F_w$ in the GFM, as pseudo-glosses provide a more meaningful representation by aligning with target semantics at a local level. 
The resulting integrated feature for diffusion conditioning, $F_{pv} = GF(F_p\ \copyright\ F_v) \in \mathbb{R}^{B \times (L_p + L_v) \times D}$, is then obtained through the GFM. Note that we use $F_{pv}$ instead of $F_{wv}$ for DiffSLT-P in subsequent subsections.

\noindent\textbf{Denoising Network.}
\label{sec:diffusion_model}
As the main component of DiffSLT, a latent diffusion model is employed to serve as the sign language translator.
Leveraging its probabilistic generative nature, DiffSLT produces diverse translations that preserve the semantics of the input sign language. This flexibility allows for exploring multiple plausible spoken sentence outputs, resulting in more natural and varied translations.

Given an input video $X$ and a target sentence $Y$, we obtain the conditioning feature $F_{wv}$ via the pretrained video feature extractor with the GFM, and the compressed latent feature $z_0$ using the compression network. 
These features serve as inputs to train the denoising network.
After sampling a timestep $t \leq T$, where $T$ is the final timestep, the noised latent feature $z_t$ is generated from the initial latent $z_0$ according to Eq.~\ref{eq:prelim_dm1}, and then passed into the denoising network $z_{\theta}$. 
In the denoising network, $z_t$ is linearly projected through a feed-forward layer, followed by the summation of learnable time and positional embeddings.
The latent is then processed through $N$ attention blocks, each consisting of self-attention, cross-attention, and feed-forward subblocks. 
In the cross-attention subblock, conditioning feature $F_{wv}$ serves as the key and value for attention. 
Finally, the output is projected into the denoised latent feature $\tilde{z}_0$.

Additionally, we employ the self-conditioning technique~\cite{chen2022analog} to better regularize to the generated latents~\cite{lovelace2024latent, yuan2022seqdiffuseq}. 
This technique leverages the model's own previous prediction $\tilde{z}_0^k$ from an earlier timestep $k$ (where $k > t$). 
$\tilde{z}_0^k$ is concatenated with the current latent feature, $z_t$, along the feature dimension, thereby conditioning the current denoising step on previous predictions.
In cases where no prior prediction is available, such as at the final timestep, a predefined initial self-conditioning embedding is concatenated with the input. 

Thus, the denoising network $z_{\theta}$ is trained to minimize the diffusion loss, $\mathcal{L}_{diff}$ as follows:
\begin{equation}
\mathcal{L}_{diff} = \mathbb{E}_{t, z} \left[ \lambda_t \| z_\theta(z_t, \tilde{z}_0^k, F_{wv}, t) - z_0 \|_1 \right].
\end{equation}
 
\noindent We apply L1 loss as our objective function following findings from prior research~\cite{lovelace2024latent} in sequence-to-sequence text generation.
During training, the LLM encoder-decoder ($\Psi_E$ and $\Psi_D$) and the visual feature extractors ($W$ and $V$) remain frozen.

\subsection{Translation Inference}
\label{sec:inference}
In the inference phase, we initialize the random latents \(z_T \in \mathbb{R}^{B \times l \times d}\) from a normal distribution, rather than using target sentence embeddings. 
The latent is denoised by $z_{\theta}$, conditioned on $F_{wv}$.
To improve sampling quality, we employ classifier-free guidance (CFG)~\cite{ho2022classifier}, which jointly reflects the conditional and unconditional denoising processes given CFG scale $w_{\text{CFG}}$:
\begin{equation}
\tilde{z}_0 = w_\text{CFG} \cdot z_\theta(z_t,\tilde{z}_0^k, F_{wv},t) + (1-w_\text{CFG}) \cdot z_\theta(z_t, t).
\end{equation}
The sampled latent feature $\tilde{z}_0$ is then reconstructed into a high-dimensional text embedding using the pretrained reconstruction network $RN$.
Finally, multiple candidate spoken sentences are generated through the LLM decoder $\Psi_D$.
Finally, we use Minimum Bayes Risk (MBR) decoding~\cite{goel2000minimum, kumar2004minimum} to select the most suitable translation by minimizing the risk associated with the loss function $\mathcal{L}_\text{MBR}$ among the candidate sentences.
The final candidate with the minimum $\mathcal{L}_\text{MBR}$ is selected, as defined below:

\begin{equation}\label{eq:mbr_loss}
\hat{y} = \arg\min_{y \in \mathcal{Y}} \sum\nolimits_{\tilde y \in \mathcal{Y}} \frac{1}{|\mathcal{Y}|} \mathcal{L}_\text{MBR}(y, \tilde{y}) ,
\end{equation}
where $y$ and $\tilde{y}$ denote one sentence and the other sentences in
the set of candidate translations, $\mathcal{Y}$, respectively, and
\(\hat{y}\) is the selected optimal translation.
$\mathcal{L}_\text{MBR}$ is a loss function defined as the negative BLEU-4~\cite{papineni2002bleu} score between two sentences, which does not contribute to the training process.

\definecolor{darklavender}{RGB}{100, 70, 120}

\section{Experiments}

\begin{table*}[t]
  \centering
  \resizebox{.9\textwidth}{!}{\setlength{\tabcolsep}{10pt} 
  \begin{tabular}{lccccc ccccc}
    \toprule
    & \multicolumn{5}{c}{Dev} & \multicolumn{5}{c}{Test} \\
    \cmidrule(lr){2-6} \cmidrule(lr){7-11}
    Method & ROUGE & BLEU-1 & BLEU-2 & BLEU-3 & BLEU-4 & ROUGE & BLEU-1 & BLEU-2 & BLEU-3 & BLEU-4 \\
    \cmidrule{1-11} 
    \multicolumn{11}{l}{\textit{Gloss-based method}} \\
    \midrule
    SimulSLT~\cite{yin2021simulslt}         & 49.21 & 47.76 & 35.33 & 27.85 & 22.85 & 49.23 & 48.23 & 35.59 & 28.04 & 23.14   \\
    STN-SLT~\cite{voskou2021stochastic}     & -     & 49.12 & 36.29 & 28.34 & 23.23 & -     & 48.61 & 35.97 & 28.37 & 23.65   \\
    PET~\cite{jin2022prior}                 & -     & -     & -     & -     & -     & 49.97 & 49.54 & 37.19 & 29.30 & 24.02   \\
    MMTLB~\cite{chen2022simple}             & 53.10 & 53.95 & 41.12 & 33.14 & 27.61 & 52.65 & 53.97 & 41.75 & 33.84 & 28.39   \\
    SLTUNET~\cite{zhang2023sltunet}         & 52.23 & -     & -     & -     & 27.87 & 52.11 & 52.92 & 41.76 & 33.99 & 28.47   \\
    TS-SLT~\cite{chen2022two}               & \textbf{54.08} & \textbf{54.32} & \textbf{41.99} & \textbf{34.15} & \textbf{28.66} & \textbf{53.48} & \textbf{54.90} & \textbf{42.43} & \textbf{34.46} & \textbf{28.95}   \\
    \cmidrule{1-11} 

    \multicolumn{11}{l}{\textit{Weakly gloss-free method}} \\
    \midrule
    TSPNet~\cite{li2020tspnet}      & -     & -     & -     & -     & -     & 34.96 & 36.10 & 23.12 & 16.88 & 13.41  \\
    GASLT~\cite{yin2023gloss}       & -     & -     & -     & -     & -     & 39.86 & 39.07 & {26.74} & {21.86} & 15.74  \\
    ConSLT~\cite{fu2023token}       & 47.74 & -     & -     & -     & 21.11 & 47.69 & - & - & - & 21.59  \\
    VAP~\cite{jiao2025visual}       & {51.47} & \textbf{52.78} & -     & -     & {26.62} & {51.28} & \textbf{53.07} & -     & -     & {26.16}  \\\rowcolor{yellow!20}
    DiffSLT-P (Ours)   & \textbf{52.65} & {51.40} & \textbf{40.05} & \textbf{32.32} & \textbf{26.95} & \textbf{52.32} & {51.90} & \textbf{40.39} & \textbf{32.38} & \textbf{26.66}  \\\rowcolor{yellow!20}
    \textcolor{darklavender}{DiffSLT-P (Oracle)} & \textcolor{darklavender}{54.93} & \textcolor{darklavender}{53.19} & \textcolor{darklavender}{42.10} & \textcolor{darklavender}{34.30} & \textcolor{darklavender}{28.74} & \textcolor{darklavender}{54.81} & \textcolor{darklavender}{53.94} & \textcolor{darklavender}{42.42} & \textcolor{darklavender}{34.24} & \textcolor{darklavender}{28.24} \\

    \cmidrule{1-11} 
    
    \multicolumn{11}{l}{\textit{Gloss-free method}} \\
    \midrule
    NSLT~\cite{camgoz2018neural}        & 31.00 & 28.10 & 16.81 & 11.82 & 9.12  & 29.70 & 27.10 & 15.61 & 10.82 & 8.35   \\
    SLRT$^\ddag$~\cite{camgoz2020sign}         & -     & -     & -     & -     & -     & 31.10 & 30.88 & 18.57 & 13.12 & 10.19  \\
    CSGCR~\cite{zhao2021conditional}    & 38.96 & 35.85 & 24.77 & 18.65 & 15.08 & 38.85 & 36.71 & 25.40 & 18.86 & 15.18  \\
    GFSLT-VLP~\cite{zhou2023gloss}          & 43.72 & 44.08 & 33.56 & 26.74 & 22.12 & 42.49 & 43.71 & 33.18 & 26.11 & 21.44  \\
    Sign2GPT~\cite{wong2024sign2gpt}    & -     & -     & -     & -     & -     & 48.90 & 49.54 & 35.96 & 28.83 & 22.52  \\
    Fla-LLM~\cite{chen2024factorized}   & - & - & - & - & - & 45.27 & 46.29 & 35.33 & 28.03 & 23.09   \\
    SignLLM~\cite{gong2024llms}     & 47.23 & 46.88 & 36.59 & 29.91 & \textbf{25.25} & 44.49 & 45.21 & 34.78 & 28.05 & 23.40   \\
    \rowcolor{yellow!20}DiffSLT (Ours)         & \textbf{50.53} & \textbf{50.15} & \textbf{38.27} & \textbf{30.38} & 24.94 & \textbf{50.80} & \textbf{50.54} & \textbf{39.13} & \textbf{31.40}  & \textbf{25.94}  \\
    \rowcolor{yellow!20}\textcolor{darklavender}{DiffSLT (Oracle)} & \textcolor{darklavender}{52.84} & \textcolor{darklavender}{52.10} & \textcolor{darklavender}{40.28} & \textcolor{darklavender}{32.21} & \textcolor{darklavender}{26.52} & \textcolor{darklavender}{52.73} & \textcolor{darklavender}{52.33} & \textcolor{darklavender}{40.86} & \textcolor{darklavender}{32.90} & \textcolor{darklavender}{27.16} \\
    \bottomrule
  \end{tabular}}
    \vspace{-.2cm}
  \caption{Quantitative results for PHOENIX14T dataset. $\ddag$ denotes results reproduced by~\cite{yin2023gloss}. 
  The highest values are highlighted in \textbf{bold}. 
  }
  \label{tab:phoenix}
\end{table*}

\begin{table*}[t]
  \centering
  \resizebox{.9\textwidth}{!}{\setlength{\tabcolsep}{10pt} 
  \begin{tabular}{lccccc ccccc}
    \toprule
    & \multicolumn{5}{c}{Dev} & \multicolumn{5}{c}{Test} \\
    \cmidrule(lr){2-6} \cmidrule(lr){7-11}
    Method & ROUGE & BLEU-1 & BLEU-2 & BLEU-3 & BLEU-4 & ROUGE & BLEU-1 & BLEU-2 & BLEU-3 & BLEU-4 \\
    \cmidrule{1-11} 
    
    \multicolumn{11}{l}{\textit{Gloss-based method}} \\
    \midrule
    SLRT*~\cite{camgoz2020sign}             & 37.96 & 37.47 & 24.67 & 16.86 & 11.88 & 36.74 & 37.38 & 24.36 & 16.55 & 11.79   \\
    MMTLB~\cite{chen2022simple}             & 53.38 & 53.81 & 40.84 & 31.29 & 24.42 & 53.25 & 53.31 & 40.41 & 30.87 & 23.92   \\
    SLTUNET~\cite{zhang2023sltunet}         & 53.58 & -     & -     & -     & 23.99 & 54.08 & 54.98 & 41.44 & 31.84 & 25.01   \\
    TS-SLT~\cite{chen2022two}               & \textbf{55.10} & \textbf{55.21} & \textbf{42.31} & \textbf{32.71} & \textbf{25.76} & \textbf{55.72} & \textbf{55.44} & \textbf{42.59} & \textbf{32.87} & \textbf{25.79}   \\
    \cmidrule{1-11} 

    \multicolumn{11}{l}{\textit{Weakly gloss-free method}} \\
    \midrule
    TSPNet$^\ddag$~\cite{li2020tspnet}      & -     & -     & -     & -     & -     & 18.38 & 17.09 & 8.98 & 5.07 & 2.97  \\
    GASLT~\cite{yin2023gloss}       & -     & -     & -     & -     & -     & 20.35 & 19.90 & 9.94 & 5.98 & 4.07  \\
    ConSLT~\cite{fu2023token}       & 41.46 & -     & -     & -     & 14.80 & 40.98 & -     & -     & -     & 14.53  \\
    VAP~\cite{jiao2025visual}       & 48.72 & 50.41 & -     & -     & 21.16 & 48.56 & 49.99 & -     & -     & 20.85  \\
    \rowcolor{yellow!20}DiffSLT-P (Ours)    & \textbf{55.71} & \textbf{54.42} & \textbf{40.78} & \textbf{30.81} & \textbf{23.83} & \textbf{55.58} & \textbf{54.31} & \textbf{40.80} & \textbf{30.86} & \textbf{23.85}  \\
    \rowcolor{yellow!20}\textcolor{darklavender}{DiffSLT-P (Oracle)} & \textcolor{darklavender}{58.30} & \textcolor{darklavender}{56.39} & \textcolor{darklavender}{43.09} & \textcolor{darklavender}{33.25} & \textcolor{darklavender}{26.12} & \textcolor{darklavender}{58.22} & \textcolor{darklavender}{56.34} & \textcolor{darklavender}{43.11} & \textcolor{darklavender}{33.24} & \textcolor{darklavender}{26.09} \\

    \cmidrule{1-11} 
    
    \multicolumn{11}{l}{\textit{Gloss-free method}} \\
    \midrule
    SLRT$^\dag$~\cite{camgoz2020sign}         & 20.51 & 21.03 & 9.97 & 5.96 & 4.04 & 19.67 & 20.00 & 9.11 & 4.93 & 3.03  \\
    NSLT+Luong*~\cite{camgoz2018neural,luong2015effective}
                                        & 34.28 & 34.22 & 19.72 & 12.24 & 7.96 & 34.54 & 34.16 & 19.57 & 11.84 & 7.56   \\
    GFSLT-VLP~\cite{zhou2023gloss}          & 36.70 & 39.20 & 25.02 & 16.35 & 11.07 & 36.44 & 39.37 & 24.93 & 16.26 & 11.00  \\
    Fla-LLM~\cite{chen2024factorized}   & -     & -     & -     & -     & -     & 37.25 & 37.13 & 25.12 & 18.38 & 14.20   \\
    Sign2GPT~\cite{wong2024sign2gpt}    & -     & -     & -     & -     & -     & 42.36 & 41.75 & 28.73 & 20.60 & 15.40  \\
    SignLLM~\cite{gong2024llms}         & 39.18 & 42.45 & 26.88 & 17.90 & 12.23 & 39.91 & 39.55 & 28.13 & 20.07 & 15.75   \\
    \rowcolor{yellow!20}DiffSLT (Ours)& \textbf{52.92} & \textbf{52.28} & \textbf{38.66} & \textbf{28.83} & \textbf{22.02} & \textbf{53.44} & \textbf{52.87} & \textbf{39.00} & \textbf{28.91} & \textbf{21.88}  \\
    \rowcolor{yellow!20}\textcolor{darklavender}{DiffSLT (Oracle)} & \textcolor{darklavender}{55.53} & \textcolor{darklavender}{54.38} & \textcolor{darklavender}{40.94} & \textcolor{darklavender}{31.04} & \textcolor{darklavender}{23.99} & \textcolor{darklavender}{55.99} & \textcolor{darklavender}{54.71} & \textcolor{darklavender}{41.22} & \textcolor{darklavender}{31.29} & \textcolor{darklavender}{24.16} \\
    
    \bottomrule
  \end{tabular}}
    \vspace{-.2cm}
  \caption{Quantitative results for CSL-Daily dataset. 
  Results reproduced by~\cite{zhou2021improving},~\cite{zhou2023gloss}, and~\cite{yin2023gloss} are denoted by *, $\dag$, and $\ddag$, respectively.
  Note that we report the evaluation results of VAP~\cite{jiao2025visual} without punctuation preprocessing~\cite{min2023faithful}.
  }
  \label{tab:csl}
    \vspace{-.4cm}
\end{table*}

\subsection{Experimental setup}
\label{sec:Experimental setup}
\noindent\textbf{Datasets.}
We evaluate our method on RWTH-PHOENIX-Weather 2014 T (PHOENIX14T)~\cite{Camgoz_2018_CVPR} and CSL-Daily~\cite{zhou2021improving} datasets. PHOENIX14T~\cite{Camgoz_2018_CVPR} consists of 8,257 pairs of sign video sequences and spoken sentences from German weather forecast news.
CSL-Daily~\cite{zhou2021improving} contains 20,654 pairs of Chinese sign language videos and sentences from daily scenarios, such as traveling and shopping.

\noindent\textbf{Metrics.}
To evaluate our proposed method, we use standard metrics, including ROUGE-L~\cite{lin2004rouge} and BLEU~\cite{papineni2002bleu}, consistent with previous studies~\cite{zhou2023gloss, gong2024llms, wong2024sign2gpt}.
ROUGE-L~\cite{lin2004rouge} measures the F1-score of the longest common subsequences, while BLEU~\cite{papineni2002bleu} assesses n-gram precision between predicted and ground truth sentences.
To comprehensively evaluate diversity, we use metrics such as Diversity~\cite{su2022contrastive}, Compression Ratio~\cite{shaib2024standardizing}, Homogenization~\cite{shaib2024standardizing}, Memorization~\cite{lovelace2024latent}, and BERTScore~\cite{zhang2019bertscore}.
Detailed explanations of each metric are provided in the supplementary.

\noindent\textbf{Implementation Details.}
Our implementation is based on the PyTorch~\cite{paszke2017automatic} framework.
For visual encoder, we adapt 12-layer MBart~\cite{liu2020multilingual} encoder with LoRA~\cite{hu2021lora} adapter for efficient pretraining in the first stage of our method.
For the diffusion model, we primarily follow the hyperparameters outlined in~\cite{lovelace2024latent}.
We set the initial learning rate to $2\cdot 10^{-4}$ and gradually decay it using a cosine scheduler over 150k iterations.
For MBR decoding, we sample five candidate spoken sentences using the DDIM~\cite{song2020denoising} sampler with 30 timesteps with cosine scheduling. 
We train our model with an NVIDIA A100 GPU for 48 hours.
Further details are provided in the supplementary material.

\subsection{Experimental Results}
In this section, we compare DiffSLT and DiffSLT-P with several existing SLT models to evaluate their translation accuracy and diversity across different settings: gloss-based, weakly gloss-free, and gloss-free models.

\noindent\textbf{Accurate Translation.} 
We present quantitative results for the PHOENIX14T~\cite{Camgoz_2018_CVPR} and CSL-Daily~\cite{zhou2021improving} datasets in Tab.~\ref{tab:phoenix} and Tab.~\ref{tab:csl}. 
As our method generates multiple translation candidates, we additionally report the oracle score~\cite{lovelace2024latent}, calculated by selecting the candidate with the highest BLEU-4~\cite{papineni2002bleu} score relative to the ground truth sentence. 
It represents an upper bound for model performance, indicating the ideal case where the best candidate is selected from all generated options.

DiffSLT and DiffSLT-P consistently outperform other SOTA models in both gloss-free and weakly gloss-free settings on both datasets.
In particular, DiffSLT shows impressive performance gains over SignLLM~\cite{gong2024llms}, achieving increases of 6.31 and 2.54 points in ROUGE-L~\cite{lin2004rouge} and BLEU-4 scores on the PHOENIX14T dataset, and 13.53 and 6.13 points on the CSL-Daily dataset. 
These substantial gains particularly emphasize the superior ability of our method in translating long-context sentences.
Qualitative results in Fig.~\ref{fig:Qualitative results} further support this, showcasing greater accuracy compared to other models.
We attribute this advantage to the diffusion model's ability to generate entire sentences, in contrast to other SLT models, which typically generate sentences progressively in an autoregressive manner.
In addition, the oracle scores of DiffSLT and DiffSLT-P are even higher than those of MBR decoding, suggesting that careful selection of the optimal candidate could further enhance performance.

\definecolor{highlightgreen}{HTML}{D9F2D0}
\definecolor{highlightred}{HTML}{FFD1D1}

\definecolor{red}{HTML}{FFD1D1}
\definecolor{green}{HTML}{D9F2D0}

\DeclareRobustCommand{\hlred}[1]{{\sethlcolor{red}\hl{#1}}}
\DeclareRobustCommand{\hlgreen}[1]{{\sethlcolor{green}\hl{#1}}}

\begin{figure}[!t]
    \centering
    \includegraphics[width=\linewidth]{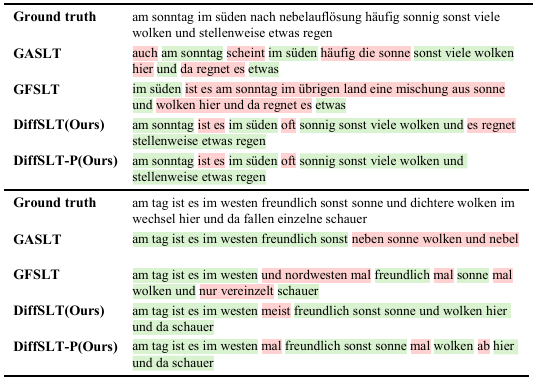}
    \caption{Qualitative results for long-context sentences from the test set of PHOENIX14T. Incorrect translations are highlighted in \hlred{red}, while accurate translations are highlighted in \hlgreen{green}.}
    \label{fig:Qualitative results}
    \vspace{-.4cm}
\end{figure}

\begin{figure}[!t]
    \centering
    \includegraphics[width=0.45\textwidth]{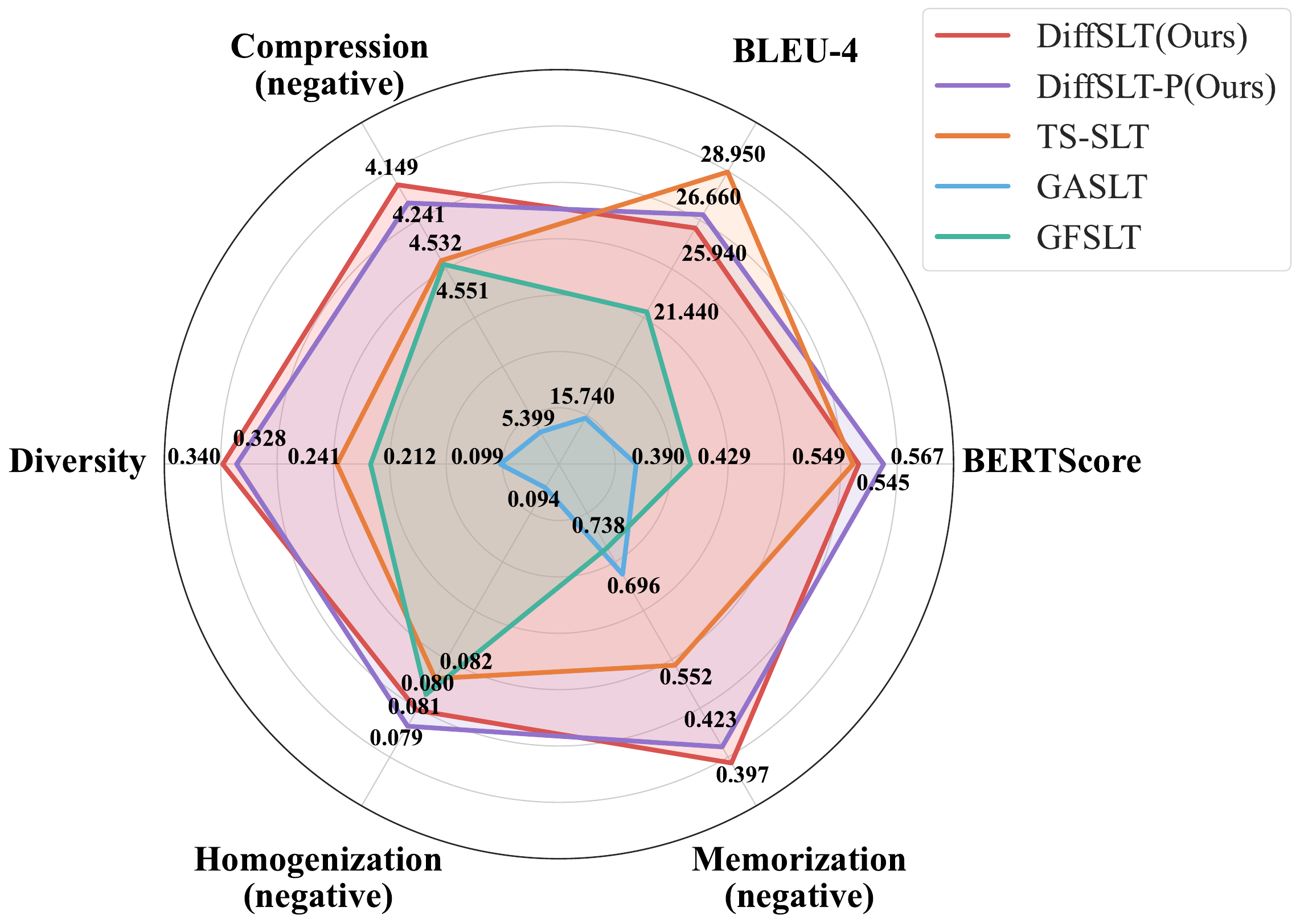}
    \caption{Evaluation results on diversity. We report the diversity metrics and BLEU-4 compared to previous SLT models across gloss-based (TS-SLT), weakly gloss-free (GASLT) and gloss-free (GFSLT) models. 
    The performance of DiffSLT and other models across multiple metrics, with higher values indicating better performance for positive metrics and lower values preferred for negative metrics.
    A model's performance is visually represented by the area covered by its polygon, where a larger area extending closer to the outermost circle signifies superior overall performance.
    }
    \label{fig:diversity results}
\end{figure}
\noindent\textbf{Diverse Translation.}
To comprehensively assess translation diversity, we evaluate DiffSLT across multiple metrics.
As shown in Fig.~\ref{fig:diversity results}, DiffSLT outperforms other SLT models across all diversity metrics, including Diversity~\cite{su2022contrastive}, Compression Ratio~\cite{shaib2024standardizing}, Homogenization~\cite{shaib2024standardizing}, Memorization~\cite{lovelace2024latent}, and BERTScore~\cite{zhang2019bertscore}. 
Our gloss-free model, DiffSLT, even outperforms the gloss-based method TS-SLT~\cite{chen2022two} in Diversity by over 36\%, highlighting DiffSLT's ability to generate varied expressions rather than fixed common phrases.
Combined with a high BERTScore, which reflects strong semantic accuracy, our methods demonstrate an impressive balance between diversity and contextual fidelity, thereby enhancing overall translation quality.
In DiffSLT-P, there is a trade-off: slightly reduced diversity but improved accuracy compared to DiffSLT. 
This may be attributed to the use of pseudo-glosses, which provide key textual guidance to the generative diffusion model, helping to reflect the syntactic information of sign language better while acting as a constraint on diversity.

\subsection{Ablation Studies}
To examine the impact of hyperparameters in the diffusion process—such as classifier-free guidance, samplers, self-conditioning, and MBR candidates—and to demonstrate the effectiveness of the proposed GFM, we conduct ablation studies on these components.
All experiments are performed using our gloss-free model on the PHOENIX14T test set.
Additional details and further ablation studies are provided in the supplementary.

\begin{figure}[t]
  \centering
    \centering
    \includegraphics[width=0.47\textwidth]{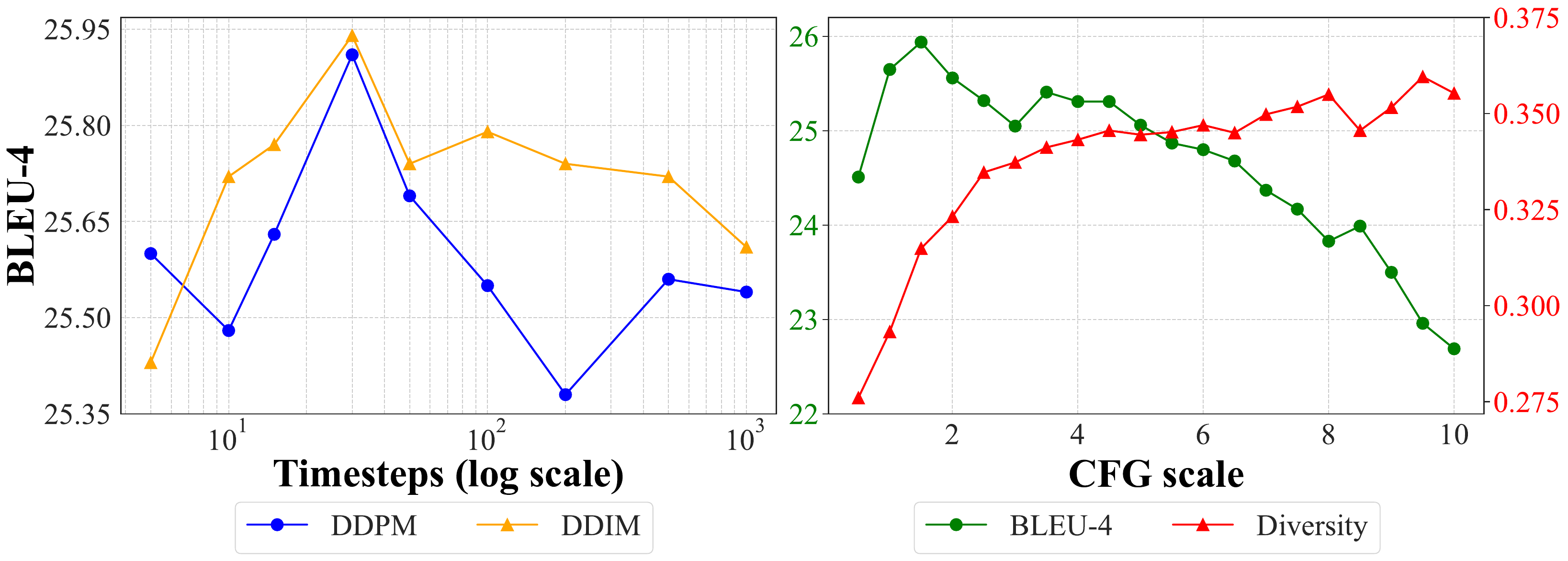}
   \caption{Sampling strategy. We validate our sampling strategy with different CFG scales and timesteps. We use DDPM~\cite{ho2020denoising} and DDIM~\cite{song2020denoising} as our diffusion samplers.
   }
   \label{fig:sampling_strategy}
\end{figure}

\noindent\textbf{Sampling Strategy.}
We conduct ablation studies on the diffusion samplers with varying timesteps and CFG scales, significantly controlling generation quality and conditioning strength. The results are presented in Fig.~\ref{fig:sampling_strategy}.
In diffusion models, more timesteps generally promote stable convergence and lead to longer inference time~\cite{rombach2022high, lovelace2024latent, yuan2022seqdiffuseq, li2022diffusion}.
However, we observe that common configurations, such as the DDPM~\cite{ho2020denoising} sampler with 1,000 steps, do not yield optimal performance in our framework.
In contrast, the overall performance with the DDIM~\cite{song2020denoising} sampler surpasses that of DDPM, achieving better BLEU-4 scores. 
Notably, using only 30 steps with DDIM yields the best results, significantly improving inference speed. 
Regarding the CFG scale, we observe that BLEU-4 scores peak at a scale of 1.5 and then gradually decline as the scale increases. 
This may indicate mode collapse at higher CFG scales~\cite{chung2024cfg++}, where conditioning features are less effectively utilized.
In contrast, Diversity steadily increases with a higher CFG scale, showing an inverse relationship with BLEU-4 scores.
This result indicates a tradeoff: strictly conditioning-dependent generation may improve translation accuracy at the cost of reduced diversity.

\begin{table}
  \centering
  \resizebox{.7\linewidth}{!}{\setlength\tabcolsep{15.0pt}\begin{tabular}{@{}lcc@{}}
    \toprule
    Method & ROUGE & BLEU-4 \\
    \midrule
    w.o. self-conditioning              & 49.81 & 24.66 \\
    self-conditioning & \textbf{50.80} & \textbf{25.94} \\
    \bottomrule
  \end{tabular}}
  \caption{Ablation study on self-conditioning. The probability is fixed at 0.5 during training.}
  \label{tab:selfcondition}
\end{table}
\noindent\textbf{Self-conditioning.}
We also perform an ablation study to evaluate the impact of self-conditioning, as shown in Tab.~\ref{tab:selfcondition}.
The results show a notable performance increase with the use of self-conditioning. 
This gain can be attributed to the stabilization and progressive refinement of latent features, enhancing the consistency of the generated outputs.

\begin{table}
\centering\resizebox{.8\linewidth}{!}{\setlength\tabcolsep{20.0pt}
  \centering
\begin{tabular}{@{}cccc@{}}
  \toprule
  Method & \# Candidates & ROUGE & BLEU-4 \\
  \cmidrule{1-4}
  \multirow{4}{*}{MBR}       & 3   & 50.53 & 25.64    \\
                              & 5   & 50.80 & \textbf{25.94}    \\
                              & 10  & 50.56 & 25.79    \\
                              & 20  & \textbf{50.88} & 25.74    \\
  \cmidrule{1-4}
  \multirow{4}{*}{Oracle}     & 3   & 52.14 & 26.76    \\
                              & 5   & 52.73 & 27.16    \\
                              & 10  & 53.49 & 27.91    \\
                              & 20  & \textbf{54.27} & \textbf{28.26}    \\
  \bottomrule
\end{tabular}
  }
  \caption{Ablation study on the number of candidates based on MBR and oracle sampling.}
  \label{tab:abl_mbr_candidates}
    \vspace{-.3cm}
\end{table}

\noindent\textbf{Number of Candidates.}
Since the quality of the best candidate in MBR decoding depends on the number of candidates~\cite{gong2022diffuseq}, we conduct an ablation study by varying this number. The results are presented in Tab.~\ref{tab:abl_mbr_candidates}, along with additional results for oracle sampling for comparison.
The performance with oracle sampling improves proportionally with the number of candidates, whereas MBR sampling achieves the best BLEU-4 score with 5 candidates.
The results indicate that increasing the number of candidates does not necessarily lead to improve performance because MBR decoding focuses on minimizing risk relative to other candidates.
In contrast, oracle sampling always selects the best option, leading to improved performance as the number of candidates increases.

\begin{table}
  \centering
  \resizebox{.7\linewidth}{!}{\begin{tabular}{@{}cccc@{}}
    \toprule
    Frame features & Video features & ROUGE & BLEU-4 \\
    \midrule
    \ding{51}  & \ding{55}              & 45.59 & 20.22 \\
    \ding{55}            & \ding{51}    & 49.58 & 24.94 \\
    \ding{51}  & \ding{51}    & \textbf{50.80} & \textbf{25.94} \\
    \bottomrule
  \end{tabular}
  }
  \caption{Impact of integrating different levels of visual features in the Guidance Fusion Module.}
  \label{tab:abl_visual_features}
\end{table}

\begin{table}
\centering\resizebox{.8\linewidth}{!}{\setlength\tabcolsep{7.0pt}
  \centering
  \begin{tabular}{@{}ccccc@{}}
    \toprule
    \# Layers & Skip connection & Early fusion & ROUGE & BLEU-4 \\
    \midrule
    1 & \ding{51}  & \ding{51}             & 49.82 & 24.51 \\
    2 & \ding{51}  & \ding{51}             & 49.95 & 24.61 \\
    3 & \ding{55} & \ding{51}              & 49.87 & 25.27 \\
    3 & \ding{51} & \ding{55}              & 47.71 & 23.56 \\
    3 & \ding{51}  & \ding{51}             & \textbf{50.80} & \textbf{25.94} \\
    \bottomrule
  \end{tabular}
  }
  \caption{
  Impact of architectural subcomponents of the Guidance Fusion Module.
  }
  \label{tab:abl_guidance_fusion_module}
    \vspace{-.2cm}
\end{table}

\noindent\textbf{Guidance Fusion Module.}
First, to evaluate the impact of integrating different levels of visual features, $F_w$ and $F_v$, in the GFM, we incrementally add both features as inputs.
As shown in Tab.~\ref{tab:abl_visual_features}, combining both visual features improves visual conditioning compared to using a single feature level.
We infer that this enhancement arises from the complementary interaction between the frame-level local information captured by $F_w$ and the global spatiotemporal context provided by $F_v$.
In addition, to investigate the impact of the module design in the GFM, we conduct experiments by varying the use of early fusion, skip connections, and the number of feed-forward layers.
As shown in Tab.~\ref{tab:abl_guidance_fusion_module}, the best performance is achieved with a three-layer feed-forward structure that integrates both skip connections and early fusion.
Notably, removing early fusion, which involves the late fusion of individually processed $F_v$ and $F_w$ through feed-forward networks, significantly decreases performance.
We conclude that for effective diffusion conditioning, it is essential to properly integrate different levels of visual features $F_v$ and $F_w$ by projecting them into a unified latent space.

\section{Conclusion}
This work emphasizes the importance of diverse translations in SLT and introduces the diffusion model as a novel solution.
We propose DiffSLT and DiffSLT-P, gloss-free and weakly gloss-free SLT frameworks, which provide accurate and diverse translations, yielding high-quality outputs and significantly surpassing existing methods.
We demonstrate the impressive potential of our approach in SLT through comprehensive experiments and analyses.
We believe that our work will serve as a milestone to inspire further research in this field, ultimately contributing to enhancing the well-being of the deaf community.

{
    \small
    \bibliographystyle{ieeenat_fullname}
    \bibliography{main}
}

\clearpage
\renewcommand{\thesection}{\Alph{section}}
\renewcommand{\thetable}{\Alph{section}.\arabic{table}}
\renewcommand{\thefigure}{\Alph{section}.\arabic{figure}}
\setcounter{section}{0}
\setcounter{table}{0}
\maketitlesupplementary

\section{Additional Ablation Studies}



\paragraph{LLM Encoder-Decoder.}
We conduct an ablation study to validate our design choice for LLM encoder-decoder. 
We evaluate the results using different multilingual sequence-to-sequence models, including MT5-Base~\cite{xue2020mt5}, MBart-Large~\cite{liu2020multilingual}, and MBart-Large-MMT~\cite{liu2020multilingual}. 
As described in Tab.~\ref{tab:ablation-llm}, the MBart-Large-MMT model achieves the best performance for both ROUGE and BLEU scores. 
Since MBart-Large-MMT is fine-tuned for the machine translation task~\cite{liu2020multilingual}, it is particularly well-suited for SLT as it leverages its capability to handle cross-lingual text generation effectively.

\begin{table}[H]
  \centering
  \begin{tabular}{@{}lcc@{}}
    \toprule
    LLM backbone & ROUGE & BLEU-4 \\
    \midrule
    MT5-Base~\cite{xue2020mt5}                 & 47.88 & 22.30 \\
    MBart-Large~\cite{liu2020multilingual}     & 49.68 & 24.13 \\
    MBart-Large-MMT~\cite{liu2020multilingual} & \textbf{50.80} & \textbf{25.94} \\
    \bottomrule
  \end{tabular}
  \caption{Ablation study on LLM backbone in DiffSLT. 
  }
  \label{tab:ablation-llm}
\end{table}

\paragraph{L1 Loss vs L2 Loss.}
Since L1 loss has been shown to facilitate high-fidelity generation in diffusion models~\cite{saharia2022palette}, we adopt this loss function for DiffSLT training.
To validate our choice, we conduct an ablation study on the loss function, with the results presented in Tab.~\ref{tab:ablation-loss}.
While L2 loss provides competitive performance compared to L1 loss, L1 loss remains preferable for achieving a high BLEU-4 score, which is consistent with previous studies~\cite{saharia2022palette, lovelace2024latent}.

\begin{table}[H]
  \centering
    \begin{tabular}{@{}lcccc@{}}
    \toprule
    Loss & ROUGE & BLEU-4\\
    \midrule
    L2 Loss                  & 50.42 & 25.14 & \\
    L1 Loss & \textbf{50.80} & \textbf{25.94} &  \\
    \bottomrule
  \end{tabular}
  \caption{Ablation study on loss type. 
  }
  \label{tab:ablation-loss}
\end{table}



\setcounter{table}{0}
\setcounter{figure}{0}
\section{Metrics for Diversity}
We provide additional explanations for the diversity metrics in this section.
Diversity~\cite{su2022contrastive} is quantified as the ratio of unique n-grams to the total number of n-grams in the predicted spoken sentences $\hat{Y}$, evaluated for n-gram sizes ranging from 2 to 4.
Compression Ratio~\cite{shaib2024standardizing} is calculated by dividing the size of the concatenated predicted sentences dataset $\hat{\mathcal{D}}$ by its compressed size.
Homogenization~\cite{shaib2024standardizing} is measured using ROUGE-L~\cite{lin2004rouge}, which assesses the similarity between pairs of predicted sentences based on their longest common subsequences.
Memorization~\cite{lovelace2024latent} refers to the proportion of 4-grams in the predicted sentences that also appear in the training set.
BERTScore~\cite{zhang2019bertscore} evaluates the cosine similarity between matched tokens in the predictions and the ground truth, indicating the semantic accuracy of the predicted sentences. This metric is useful due to its flexibility in capturing semantic meaning, unlike BLEU~\cite{papineni2002bleu}, which strictly evaluates n-gram correspondences.

\begin{algorithm}[t]
\caption{DiffSLT Training Procedure}
\label{alg:diffslt_training}
\begin{algorithmic}

\STATE \textbf{Input:} Sign language video $X$ and spoken sentence $Y$
\vspace{1mm}
\STATE \textbf{Stage 1: Pretraining}
\vspace{1mm}

\STATE Define $TD(\cdot)$ as the text decoder
\WHILE{not converged}
\STATE $F_w = W(X)$; $F_v = V(F_w)$ 
\STATE Predict sentence $\hat{Y} = TD(F_v)$
\STATE Compute loss $\mathcal{L}_{vid}$ and update $W$, $V$, $TD$
\ENDWHILE
\vspace{2mm}

\STATE Freeze $\Psi_E$, $\Psi_D$
\WHILE{not converged}
\STATE $F_t = \Psi_E(Y)$
\STATE Compress $z_0 = CN(F_t)$; Reconstruct $\hat{F}_t = RN(z_0)$
\STATE Generate target sentence $\hat{Y} = \Psi_D(\hat{F}_t)$
\STATE Compute loss $\mathcal{L}_{ae}$ and update $CN$, $RN$
\ENDWHILE
\STATE \textbf{Output:} $W, V, CN, RN$
\vspace{2mm}

\STATE \textbf{Stage 2: DiffSLT Training}
\STATE Freeze $W$, $V$, $\Psi_E$, $CN$
\WHILE{not converged}
\STATE $F_t = \Psi_E(Y)$ 
\STATE $F_w = W(X)$; $F_v = V(F_w)$
\STATE $F_{wv} = GF(F_w\ \copyright\ F_v)$
\STATE Compress into latent $z_0 = CN(F_t)$
\STATE Sample $t \sim \text{Uniform}(1, T)$ and $\epsilon \sim \mathcal{N}(0, I)$
\STATE Compute $\bar{\alpha}_t = \prod_{s=1}^{t} \alpha_s$; $z_t = \sqrt{\bar{\alpha}_t}\, z_0 + \sqrt{1 - \bar{\alpha}_t}\, \epsilon$
\STATE Denoise $\tilde{z}_0 = z_\theta(z_t, \tilde{z}_0^k, F_{wv}, t)$
\STATE Compute loss $\mathcal{L}_{diff}$ and update $z_\theta$, $GF$
\ENDWHILE
\STATE \textbf{Output:} $z_\theta, GF$

\end{algorithmic}
\end{algorithm}

\setcounter{table}{0}
\setcounter{figure}{0}
\section{Overview of DiffSLT Training}
In our main manuscript, we present our strategy to pretrain the video feature extractor and the compression-reconstruction network, which facilitates training of the denoising network $z_\theta$.
We provide an overview of DiffSLT training process in Algorithm ~\ref{alg:diffslt_training}.

\setcounter{table}{0}
\setcounter{figure}{0}
\section{More Implementation Details}
\label{sec:more_implementation_details}
As described in Tab.~\ref{tab:impl_details}, we provide additional implementation details, including hyperparameters used in our method.
Here, the sampling schedule scale refers to the scale of the shifted cosine schedule~\cite{hoogeboom2023simple,chen2023importance}. 
A lower value of this scale indicates that the noise schedule is concentrated on higher noise levels, which is known to enhance the utilization of conditioning~\cite{ye2023dinoiser}.
We empirically find that our pseudo-gloss model, DiffSLT-P, requires neither as high noise levels nor as many sampling steps as the gloss-free model, DiffSLT. 
This is attributed to the pseudo-gloss which provides textual guidance, sharing the same modality as the target spoken sentence, thereby reducing the need for excessive noise or sampling steps to effectively utilize the conditioning.

\begin{table}[h!]
\footnotesize
  \centering
  \resizebox{.9\linewidth}{!}{\setlength\tabcolsep{15pt}\begin{tabular}{@{}lcc@{}}
    \toprule
     & \multicolumn{2}{c}{Method} \\
     \cmidrule(lr){2-3}
     & DiffSLT & DiffSLT-P \\
    \midrule
    Batch Size          &   \multicolumn{2}{c}{8}   \\
    Learning Rate        &  \multicolumn{2}{c}{2e-4}  \\
    Learning Rate Schedule &  \multicolumn{2}{c}{Cosine Decay} \\
    Gradient Clipping        &  \multicolumn{2}{c}{0.4}       \\
    Loss Type                    & \multicolumn{2}{c}{L1}  \\
    Optimizer               &  \multicolumn{2}{c}{AdamW~\cite{loshchilov2017decoupled}} \\
    Attention Blocks                &  \multicolumn{2}{c}{12}  \\
    Objective                    &  \multicolumn{2}{c}{$x$-prediction}         \\
    Text Embedding Dimension    & \multicolumn{2}{c}{1024} \\
    Frame Feature Dimension    & \multicolumn{2}{c}{1024} \\
    Video Feature Dimension    & \multicolumn{2}{c}{1024} \\
    Training Iteration         & \multicolumn{2}{c}{150k} \\
    Sampling Schedule          &  \multicolumn{2}{c}{Cosine Schedule} \\
    Sampling Schedule Scale     & 0.1 & 0.3 \\                       
    Sampler                     & \multicolumn{2}{c}{DDIM~\cite{song2020denoising}} \\
    Sampling Steps              & 30 & 15   \\
    CFG Scale~\cite{ho2022classifier} & \multicolumn{2}{c}{1.5}   \\
    Self-conditioning Prob.~\cite{chen2022analog} &  \multicolumn{2}{c}{0.5}    \\
    \bottomrule
  \end{tabular}}
  \caption{Implementation details of DiffSLT and DiffSLT-P.}
  \label{tab:impl_details}
  \vspace{-.4cm}
\end{table}




\setcounter{table}{0}
\setcounter{figure}{0}
\section{Number of Parameters}
In Tab.~\ref{tab:parameters-nums}, we report the number of parameters used during DiffSLT training process. 
Note that all components, except the denoising network, remain entirely frozen during DiffSLT training.

\begin{table}[h!]
\resizebox{\linewidth}{!}{\setlength\tabcolsep{4.0pt}
  \centering
  \begin{tabular}{@{}lcc@{}}
    \toprule
    Components                      & \# Parameters & \# Trainable Parameters   \\
    \midrule
    Visual Encoder                  & 637M   & 26M*               \\
    LLM Encoder-Decoder             & 611M   & 0                  \\
    Compression-Reconstruction      & 45M    & 45M*               \\
    Denoising Network                 & 419M   & 419M               \\
    \cmidrule{1-3}
    Total                           & 1,712M & 490M               \\
    \bottomrule
  \end{tabular}}
  \caption{Number of model parameters. 
  * denotes trainable parameters in pretraining.}
  \label{tab:parameters-nums}
\end{table}

\setcounter{table}{0}
\setcounter{figure}{0}
\section{Visualizations}    
\begin{figure}[h!]
  \centering
   \includegraphics[width=0.7\linewidth]{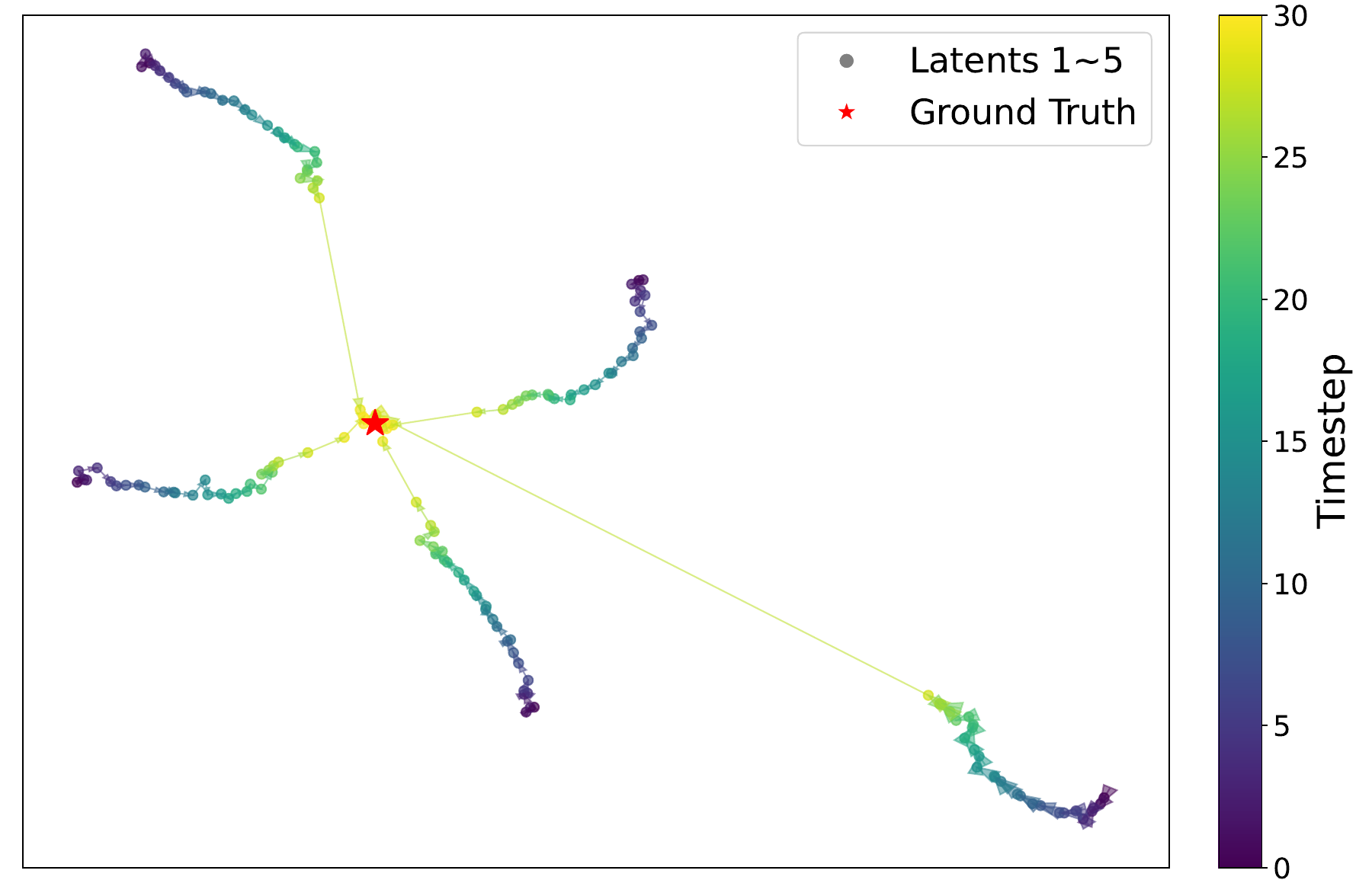}
   \vspace{-.1cm}
   
   \caption{UMAP~\cite{mcinnes2018umap} visualization of latent sampling trajectories. We visualize five sampled latents and ground truth latent across the sampling timesteps.}

   \label{fig:latent_path_umap}
   \vspace{-.8cm}
\end{figure}

\paragraph{Converging Latents.}
Fig.~\ref{fig:latent_path_umap} shows how the sampled latents converge into the target latent $z_0$ during the inference phase. 
The latents $z_t$ are randomly generated and try to reconstruct the target latent by conditioning on visual features.
We observe that these latents suddenly converge in the last few timesteps. 
This emerging convergence may be attributed to the low sampling schedule scale, which pushes most timesteps into high noise levels, as described in Sec.~\ref{sec:more_implementation_details}. 
\vspace{-.3cm}

\begin{figure}[h!]
  \centering
   \includegraphics[width=0.7\linewidth]{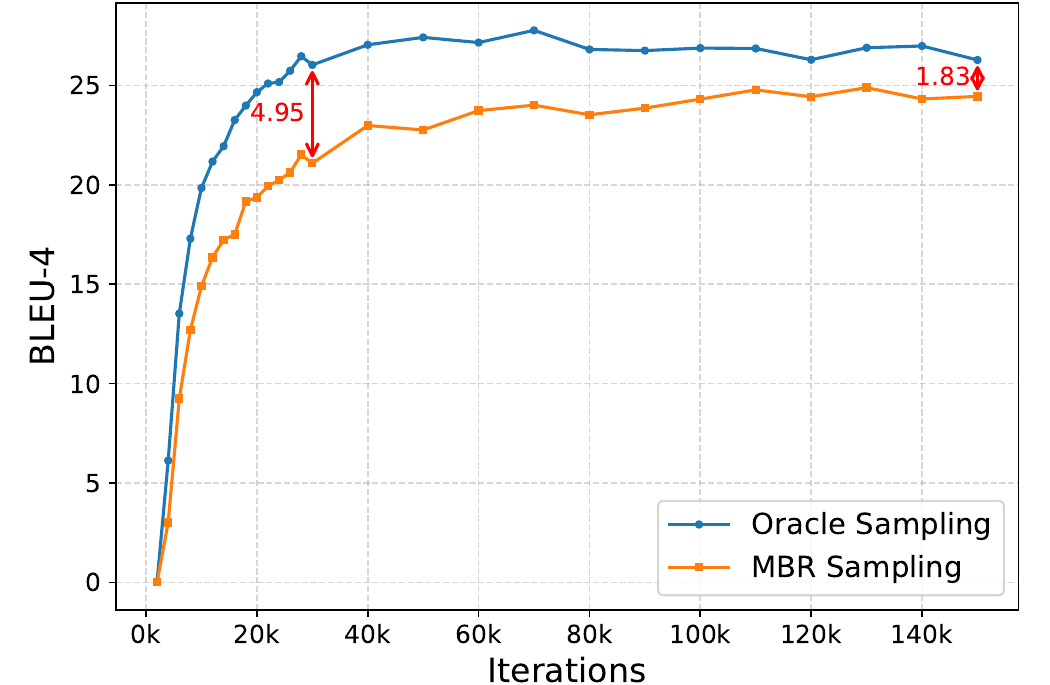}

   \vspace{-.2cm}
   \caption{BLEU-4 score in oracle sampling and MBR sampling.}
   \label{fig:vis_oracle_mbr}
   \vspace{-.5cm}
\end{figure}

\paragraph{Oracle Sampling.}
Fig.~\ref{fig:vis_oracle_mbr} shows the BLEU-4 scores of oracle sampling and MBR sampling in the training process.
Initially, there is a significant gap between oracle and MBR sampling, indicating that the quality of the generated candidates varies widely.
When candidate quality varies but has not yet stabilized at the intermediate training iteration, the disparity between oracle and MBR sampling is most pronounced. 
Over time, MBR sampling converges closely to oracle sampling, resulting in consistently high-quality generated candidates.
This occurs because oracle sampling selects the best candidate, while MBR prioritizes risk minimization, leading to safer but potentially less optimal choices when candidate quality varies.

\setcounter{table}{0}
\setcounter{figure}{0}

\setcounter{table}{0}
\setcounter{figure}{0}
\section{Additional Qualitative Results}
\begin{figure*}[!t]
    \centering
    \includegraphics[width=\textwidth]{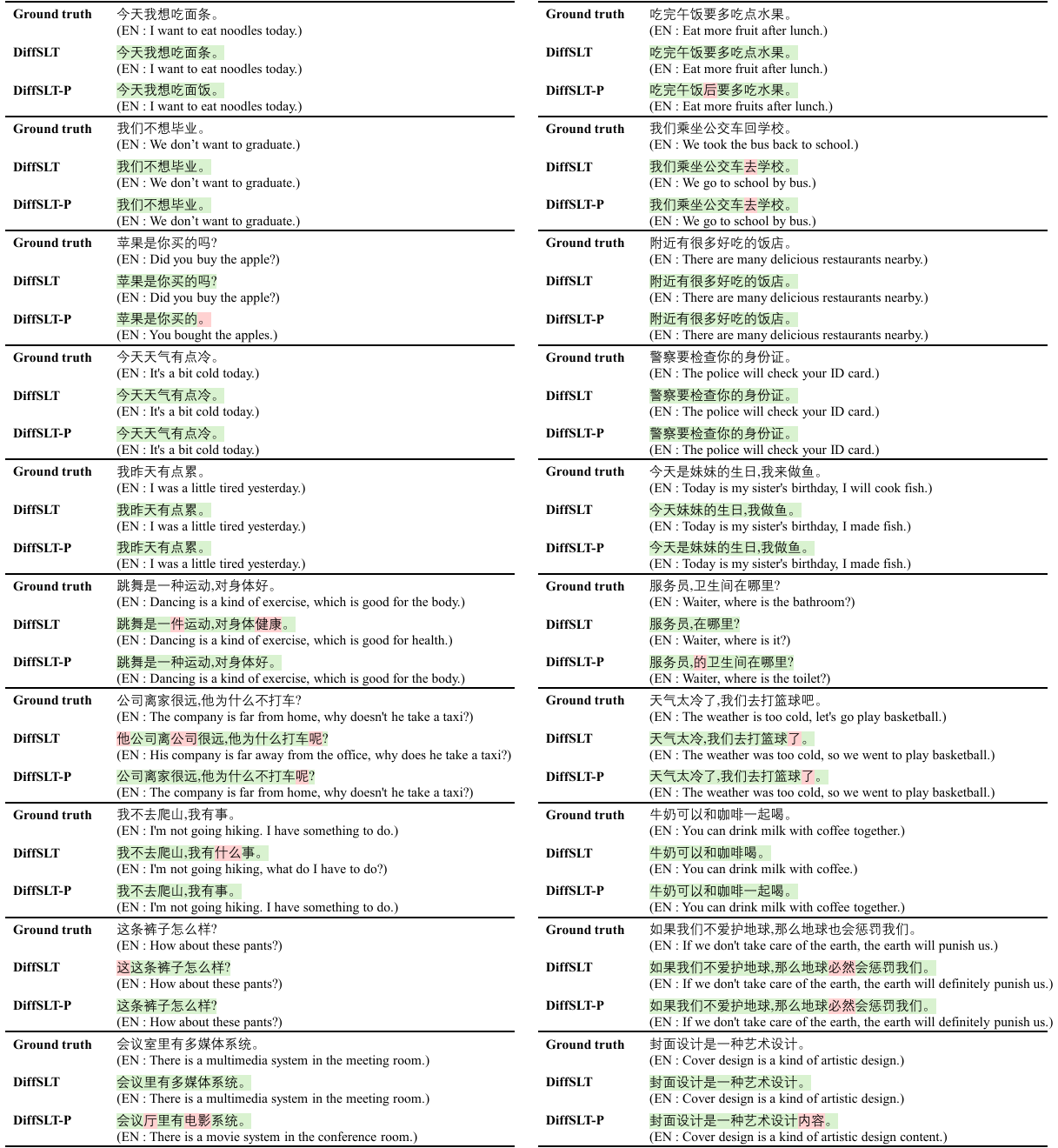}
    \caption{
        Qualitative results for test split of 
        CSL-Daily~\protect\cite{zhou2021improving} dataset.
    }
    \label{fig:qualitative_csl_daily}
\end{figure*}

\vspace{-.1cm}
We provide additional qualitative results on both datasets, PHOENIX14T~\cite{Camgoz_2018_CVPR} and CSL-Daily~\cite{zhou2021improving} in Fig.~\ref{fig:DiffSLT_part1} and Fig.~\ref{fig:qualitative_csl_daily}, respectively.
Note that GASLT~\cite{yin2023gloss} and GFSLT~\cite{zhou2023gloss} are our reproduced models. 

\definecolor{highlightgreen}{HTML}{D9F2D0}
\definecolor{highlightred}{HTML}{FFD1D1}

\definecolor{red}{HTML}{FFD1D1}
\definecolor{green}{HTML}{D9F2D0}

\DeclareRobustCommand{\hlred}[1]{{\sethlcolor{red}\hl{#1}}}
\DeclareRobustCommand{\hlgreen}[1]{{\sethlcolor{green}\hl{#1}}}

\begin{figure*}[!t]
    \centering
    \includegraphics[width=\textwidth]{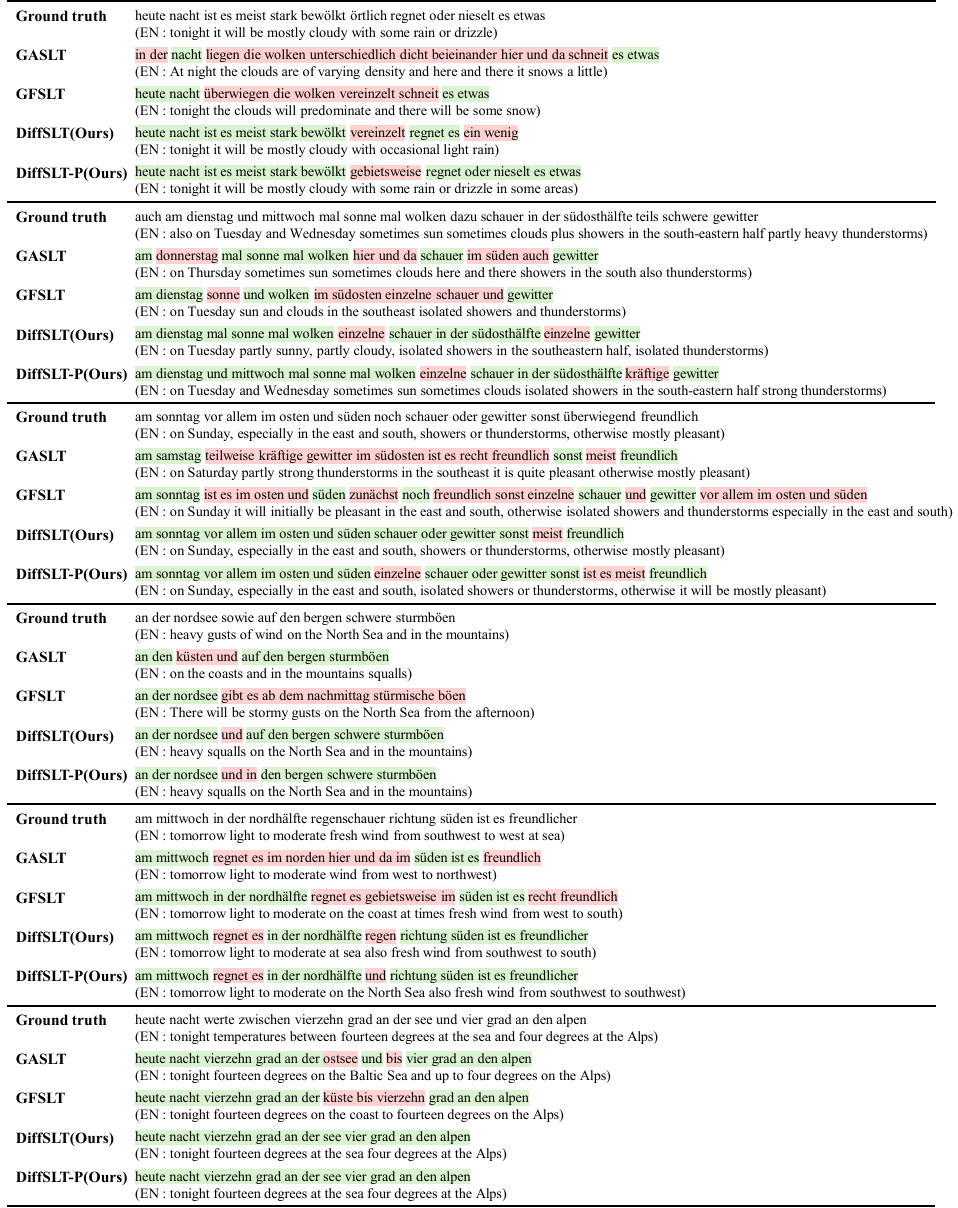}
    \caption{
        Qualitative results for test split of PHOENIX14T~\cite{Camgoz_2018_CVPR} dataset.
        Incorrect translations are highlighted in \hlred{red}, while accurate translations are highlighted in \hlgreen{green}.
    }
    \label{fig:DiffSLT_part1}
\end{figure*}

\begin{figure*}[!t]
    \ContinuedFloat
    \centering
    \includegraphics[width=\textwidth]{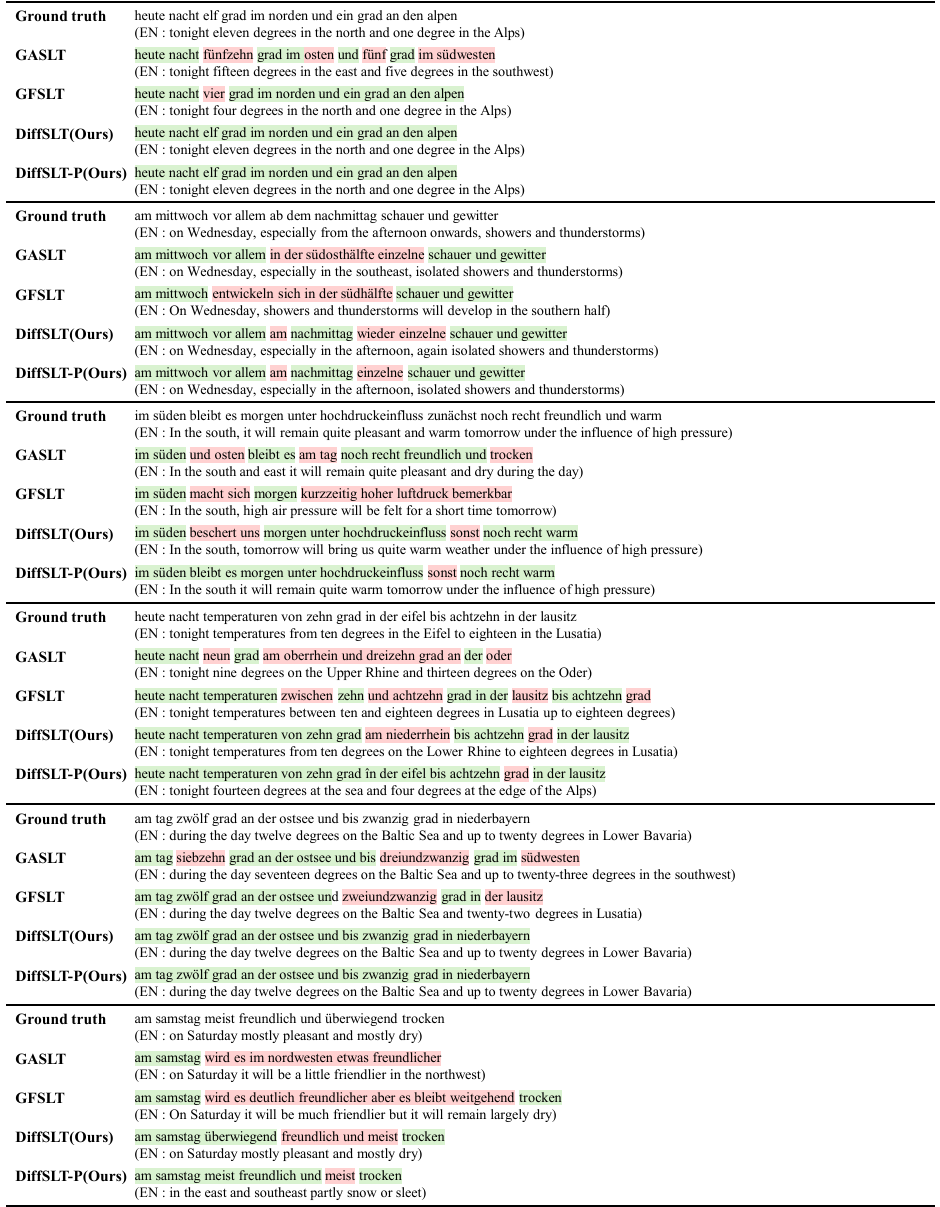}
    \caption{
        Continued: qualitative results for test split of 
        PHOENIX14T~\protect\cite{Camgoz_2018_CVPR} dataset.
    }
\end{figure*}

\end{document}